# Control and Evaluation of Event Cameras Output Sharpness via Bias

Mehdi Sefidgar Dilmaghani*[1,2] Waseem Shariff[1,2], Cian Ryan[2], Joe Lemley[2], and Peter Corcoran[1]

[1] University of Galway

[2] Xperi Corporation

## ABSTRACT

Event cameras also known as neuromorphic sensors are relatively a new technology with some privilege over the RGB cameras. The most important one is their difference in capturing the light changes in the environment, each pixel changes independently from the others when it captures a change in the environment light. To increase the user's degree of freedom in controlling the output of these cameras, such as changing the sensitivity of the sensor to light changes, controlling the number of generated events and other similar operations, the camera manufacturers usually introduce some tools to make sensor level changes in camera settings. The contribution of this research is to examine and document the effects of changing the sensor settings on the sharpness as an indicator of quality of the generated stream of event data. To have a qualitative understanding this stream of event is converted to frames, then the average image gradient magnitude as an index of the number of edges and accordingly sharpness is calculated for these frames. Five different bias settings are explained and the effect of their change in the event output is surveyed and analyzed. In addition, the operation of the event camera sensing array is explained with an analogue circuit model and the functions of the bias foundations are linked with this model.

## 1. INTRODUCTION

Event cameras also referred as neuromorphic sensors are bio-inspired imaging sensors that function very differently from traditional cameras. Instead of taking intensity images at a fixed capturing rate, each pixel in the event camera measures changes in the light intensity independently as it takes place. It is built on the notion that, like with the human eye, machines can function more efficiently if they simply absorb changes in a scene rather than constantly analyzing everything in a scene [1]. As previously explained an event camera does not capture an image frame, but rather records pixel-level changes in brightness across an imaged field-of-view (FoV). Where objects or elements within an imaged scene move within the FoV then changes will be registered related to those objects or elements. From the camera's perspective it records a change in the illumination level, either +ve or -ve, for each pixel of the camera and once a threshold change is recorded it creates an output event – a message with the XY location of that pixel, T- timestamp of event occurrence and P- polarity, whether the change is -ve (darker) or +ve (brighter). Naturally each such event on its own is relatively meaningless, but when a stream of events is grouped together, they provide an indication of a more global change in the FoV, often defining the edge or a moving object, or a pattern of change in the texture of a surface. We can change the number of collected events that are grouped for processing, in addition it is possible to adjust the parameters that cause sensor level changes and individual events to trigger, referred to as sensor biases. These include the following: bias_fo specifies the amount of background noise, bias_hpf has the role of decreasing the amount of low frequency noise which is mostly the noise generated by sensor and environment thermal noise, bias_diff_on and bias_diff_off determine the sensor sensitivity to the amount of the light change in the environment and finally the bias_refr controls the amount of the sensors sleep time between two events

Most of the recent research have focus on the potential applications of these cameras and taking the most advantage of their unique properties [2] and very few research, have studied the output characteristics of these cameras and ways to control and refine it for a particular use-cases. Changing each of the biases mentioned above, leads to a new stream of events and some analytical methods are required to decide about how to tune the output for a particular use. One way of measuring the quality of the event cameras output is to analyze the raw stream of the events which also requires proposing new metrics specific to this type of data. Even if such metrics can be very valuable and significant, it is important at this stage to have a deeper understanding of this new technology, compare its output with that of conventional RGB cameras, and use metrics that are well-known to computer vision professionals. To accomplish it, the events should be converted to a format analyzable with image quality metrics. In this paper, first we record events in different bias settings, then convert the event streams into frames and at the end we analyze the quality of these frames. As was previously mentioned, a cluster of events indicate to the edges of a moving object. Therefore, evaluating the sharpness of an event frame is a suitable technique to assess the output quality of an event camera. The metric used to measure the sharpness will be average image gradient magnitude or shortly average gradient (AG) [3].

## 2. LITERATURE REVIEW

This research gives a better understanding about the ways of controlling event cameras. There is a brief explanation about the event cameras hardware in the methodology part. To further assess the literature, the authors of [4] have explained the event camera with more technical details. In [5] the authors have explained the concept of event sensing and the physical working of the event camera, this paper further looks at the contrast threshold settings with the emphasis on the mathematics behind it, more than its circuit. Moreover, authors in [6], explain the feedback controlling of event cameras output by describing different hardware blocks of the sensor.

The main goal of this research is to find a way to evaluate the quality of event camera output with bias change. For the RGB image quality assessment (color, sharpness, contrast, noise etc.) there is a significant amount of research conducted, but there is much less research on the evaluation of the quality of event data. The main reason is that this is a new technology and there are many potential applications for these cameras which have not been investigated yet. The only factor which has been studied in detail is the background activity which is the sum of junction leakages and environment thermal noise. However, in neuromorphic cameras, there are primarily four forms of random noise. First, even though there is no true intensity change, an event is formed. These false alarms, known as "background noise or background activity" (BA), have a negative influence on algorithm performance and use bandwidth. Secondly, although there's variation in intensity, no event is created. Furthermore, the occurrence of the event is unpredictable. Finally, the series of events is proportional to the magnitude of the edge (e.g., a brightness and contrast changes creates more events than a low contrast change), the precise count of events for a given magnitude fluctuates unpredictably [4]. The majority of known event denoising algorithms focus on eliminating background activity which include both bio-inspired filtering [7], hardware-based filtering [2, 8] and by yielding spatial-temporal correlation filter [9, 10].

## 3. METHODOLOGY

Before explaining more details about the methodology, we need to briefly clarify how the event camera works first. Figure 1 shows the circuit of one pixel in an event camera. The light changes in the environment trigger the photodiode (PD) in photoreceptor, the resulting current goes through the next levels to generate an event. At the beginning there is a buffer which transfers the input current and then we have the reset circuit which puts the sensor in sleep mode as the $C_1$ and $C_2$ capacitors get charged and discharged. The last stage determines the sensitivity of the camera to positive and negative light changes with its Ion and Ioff current sources. Contrary to RGB images and videos, there are no metrics available to assess the quality of event data, and the frames produced by event cameras are frequently utilized without being properly assessed, which results in low dataset quality. On the other hand, making sensor level changes in the event camera settings named bias results in big changes in the event outputs distinguishable by naked eye, however a mathematical scheme is required to decide which set of bias amounts is the best one for each application. To the best of our knowledge, this research is the first to attempt to evaluate the sharpness of event output as the bias levels vary.

Since event cameras output indicate edge of an object or surface with brightness change, sharpness is an important factor in representing their quality. According to [3], one reliable way to assess a frame's sharpness is based on the AG. Of the several sharpness metrics, we chose this one due to its potential for quality assessment using only one image and without the necessity for comparison with a different reference image. The sharper the edges of a frame the higher value AG will be, and the manufacturers default value for most of the biases lead to such frames. So, the main question which arises is, if the default bias values are leading to the ideal output frames, what is the necessity in tweaking the camera bias. The answer is the environmental situation. These default biases have the best output in normal daylight situation with no noise, but in a different situation the same biases will fail in presenting the events. For instance, in a room with IR emissions, the event output will be noisy, and we'll need to change the bias_fo to a lower value and decrease the background noise to get a better output. The trade-offs in bias tunings should also be considered. For example, in the above application it should be considered that, lower values of bias_fo, will make the sensor slower. To achieve this first it is necessary to accumulate stream of events into frames, and second, evaluate the sharpness of that frame using AG magnitude. To examine the impact of each bias change, all the bias values are left at their factory-set default value aside from the one which is going to be assessed (test-bias). 10 different values are assigned to this test-bias and then for each value it is investigated as follows: at first, 10 seconds of head movements from left to right and vice versa is recorded, then the stream of events are accumulated to frames within constant period. Further, AG magnitude is calculated for all the accumulated frames separately, the mean value of all these magnitudes is calculated. This process will be repeated for all the values of all biases. Before discussing about the test results, more detailed information about the camera used in the tests, image gradient and biases is presented below.

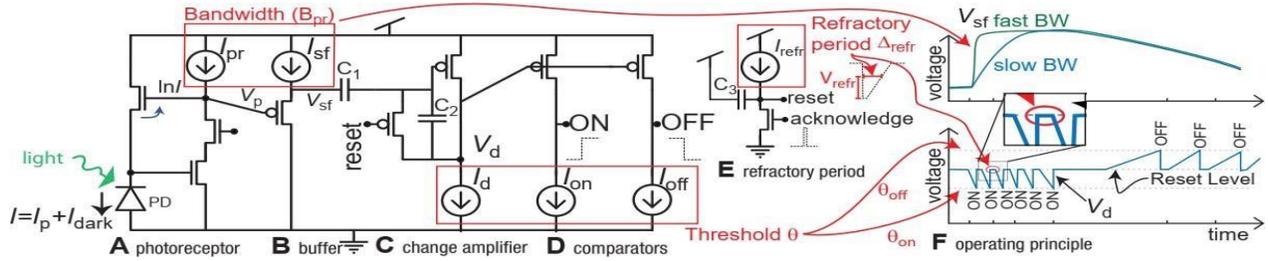

Figure 1. Circuit of each pixel in an event camera [6]

### 3.1 Event Camera Specification

The event camera utilized in this experiment is Prophesee's "Evaluation Kit-v3 - Gen4H 720P-CD" with the following configurations: resolution (px): 1280 x 720, optical format: 1/2.5", pixel latency 220 s, dynamic range > 86dB, nominal contrast threshold: 25%, pixel size: 4.86 m x 4.86 m, event signal processing embedded [11].

### 3.2 Image Gradient

Image gradients are simply directional changes in image brightness. We can identify edges in an image by estimating the gradient for a pixel of the image and continuing the procedure for the full image [12]. For the event data, since we have stream of events instead of images, as explained before, we will need to convert this stream in to frames to be able to propose an index for their edges. Image gradient computing includes determining the magnitude and direction of the gradient. Since in event cameras pixels are operating independent from each other, in this research average gradient magnitude becomes of great importance compared to the gradient direction which requires a separate study for event output. We must first determine the gradient in the x and y axes [13] then calculate the magnitude based on them and the end find the average amount of this gradient for all the pixels. Formula 1 is used to find the average amount of the image gradient magnitude or shortly the average gradient (AG):

$$Average\ Gradient\ (AG) = \frac{1}{(m-1)(n-1)} \sum_{i=1}^{m-1} \sum_{j=1}^{n-1} \sqrt{\left(\frac{\partial x(i,j)}{\partial x}\right)^2 + \left(\frac{\partial y(i,j)}{\partial y}\right)^2} \qquad (1)$$

where, m and n are the number of pixels at each frame in the x and y axes respectively. The higher the AG, the higher the number of the edges on that frame.

### 3.3 Event Camera Biases

Prophesee's event sensor provides several camera configurations that allow us to experiment with the camera performance. Different vision applications require different camera settings. As an example, tracking a driver face when he doesn't move his head a lot requires a very alert sensor and event camera manufacturers give us this degree of freedom to make such sensor level changes in their camera settings. These sensor settings are referred as bias [11] and cover different configurations from controlling the number of events generated by the camera to controlling the camera's sensitivity to positive or negative light changes in the environment.

#### 3.3.1 Buffer Bandwidth (bias_fo)

If we consider the photoreceptor circuit as a 2nd order low-pass filter, Ip together with $I_{pr}$ will set the first stage cut-off and the source follower buffer bias current $I_{sf}$ will set the second stage cut-off. This bias controls the $I_{sf}$ in the circuit shown in figure 1. By changing the amount of this current the bandwidth of the post photoreceptor source follower is being determined. Increasing the bandwidth leads to a faster sensor, however, increment in noise as well, and there is trade-off between the amount of the noise in the output and the speed of the sensor which needs to be decided according to the application specific requirements.

### 3.3.2 High-Pass Filter (bias_hpf)

Due to junction leakage currents in sensor and thermal noise, event cameras generate a noise called background activity [14] which has a low frequency. To remove this noise and other low frequency variations such as motion effects or 50 Hz lighting variation, a high pass filter is utilized. This bias controls the strength of that high pass filter. Setting it to a higher value might result in good noise removal, however it degrades the main signal as well.

### 3.3.3 Contrast Sensitivity Threshold Biases (bias_diff, bias_diff_on, bias_diff_off)

The bias_diff is a reference value and normally shouldn't be changed. The bias_diff_on and bias_diff_off control the sensitivity of the sensor to positive and negative light changes in the environment, respectively. The closer these two biases to the reference value, the more sensitive the sensor will be to the light changes. Contrast sensitivity has a linear relation with $C_2/C_1$ and $\ln(I_{on,off}/I_d)$ in the circuit shown at figure 1. This bias has shown some overlap with the bias_fo, and it should be considered when tuning these two.

### 3.3.4 Deadtime (bias_refr)

As shown in the graphs of figure 1, the sensor goes off after generating each event and doesn't follow the light changes in the environment. This period is called deadtime and is a function of $C_3/(I_{ref}*V_{ref})$. Setting it to a high value means that the sensor will be dead for a long time and less events will be generated meanwhile [11, 14]. This decrement in the number of events can be useful in the applications with low power consumption, or limited amount of accessible memory. In the applications like face tracking, where the movements are very slow and rare, the camera needs to be very alert with shorter deadtimes.

## 4. RESULT AND DISCUSSION

In this article, stream of events of a moving object are recorded and the effect of bias change on the sharpness of the output frames is studied. The test is conducted under steady conditions (considering environmental light, lens focus, etc.) throughout the whole bias ranges. It is preferred to measure the sharpness of objects suitable for potential applications like the method proposed by Ryan et al. for face detection and tracking in [15]. Due to this, the head movements are recorded instead of a penduline or any other object with sharp edges. A subject's face region is chosen as the baseline and the head is slowly moving from side-to-side to generate reference FoV. A further motivation for choosing the human face region is the potential of neuromorphic vision systems to detect detailed facial activity that can be correlated with physiological processes that can, in turn, provide a deeper understanding of the cognitive and emotional state of a human subject [16]. The human face region includes useful sub-features and a useful mix of high-contrast and low-contrast regions. This is an exploratory study and as our focus is to understand the qualitative effects of biases on a typical human face recordings that the same subject was used for all evaluations. The biases and their effect on the output quality are the focus of attention more than the subjects used in this instance. Since it was essential to have uniform illumination across all biases to have a clear understanding of the various bias settings, we used the office daylight lighting. It should be noted that altering the test situation will have an equal impact on all the biases and we will experience varying frame sharpness and AG values for each bias, but the AG curve shape will not change. For the initial study on bias variations each bias is varied individually across its recommended range of values. The remaining 4 biases are kept at their default setting. This will allow a qualitative understanding of how variations in each bias setting affect the output. The default amount, the possible ranges of each bias and their 10 different tested values are shown in Table 1.

Table 1. The range, default and tested values per each bias

| Bias | Range | Default | Tested Values | | | | | | | | | |
|---|---|---|---|---|---|---|---|---|---|---|---|---|
| | | | 1st | 2nd | 3rd | 4th | 5th | 6th | 7th | 8th | 9th | 10th |
| bias_fo | 0-255 | 74 | 0 | 15 | 30 | 45 | 60 | 74 | 90 | 105 | 120 | 135 |
| bias_hpf | 0-255 | 0 | 0 | 28 | 56 | 84 | 112 | 140 | 168 | 196 | 224 | 255 |
| bias_diff_on | 81-255 | 115 | 81 | 100 | 115 | 138 | 157 | 176 | 195 | 214 | 233 | 255 |
| bias_diff_off | 0-79 | 52 | 0 | 9 | 18 | 27 | 36 | 45 | 52 | 61 | 70 | 79 |
| bias_refr | 0-255 | 68 | 0 | 25 | 50 | 68 | 100 | 125 | 150 | 175 | 200 | 225 |

## 4.1 Bias_fo

The first bias tested is bias_fo, it's already explained that increasing this bias leads to a larger bandwidth and more signal and noise at the same time. Since the event camera output stream is converted to edges in the frames, frames with more edges and consequently, higher amounts of AG are expected for higher values of bias_fo. Figure 2 aligns with expectations and looking at the sample output frames, and the graph proves the correctness of our claim visually and mathematically. As it can be seen in the curve, the AG reaches to its maximum value for the biases greater than 100 and this will be the ideal range for this bias in our test scenario.

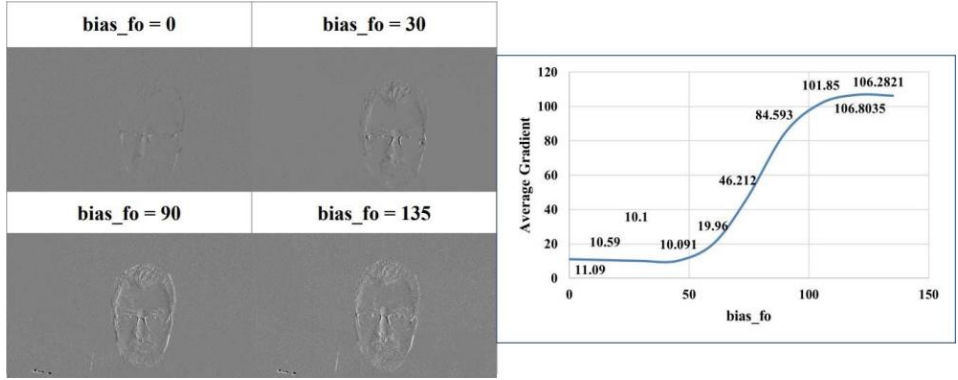

Figure 2.  Left- sample output frame per each value of bias_fo. Right- AG Vs. bias_fo change.

## 4.2 Bias_hpf

As mentioned before, the AG decreases with increment in this bias. Although at the beginning the AG shows disordered increments and decrements, all the values are in the same range, and it can't be interpreted as serious changes. When bias is increased to higher values the AG approaches to zero and this was predictable because in higher bias_hpf values, the low-speed movements can't get detected by the event camera and consequently there will be no events in the output and no edges in the generated frames. Since there are no outputs for the higher values of bias_hpf, we will depict only a few sample frames for lower values in figure 3.

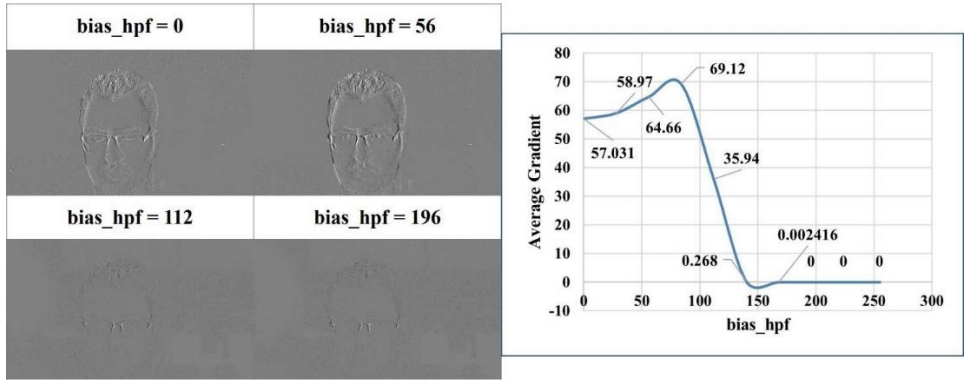

Figure 3. Left- sample output frame per each value of bias_hpf. Right- AG Vs. bias_hpf change

The graph shown confirms the theories about this high-pass filter bias and approaches to 0 after a while. The ideal range for this bias will be amounts less than 100.

## 4.3 Bias_diff_on

This bias is important in determining the number of generated events and the sensitivity of the camera to positive light changes. The more sensitive the camera, the more events will be generated. With this brief explanation it is expected to have more edges and higher AGs for higher sensitivities which occur at lower values of bias_diff_on where this bias is closer to its reference value. Figure 4 confirms this theory-based predictions in a practical scenario. As it can be seen from the curve the values around 100 are ideal.

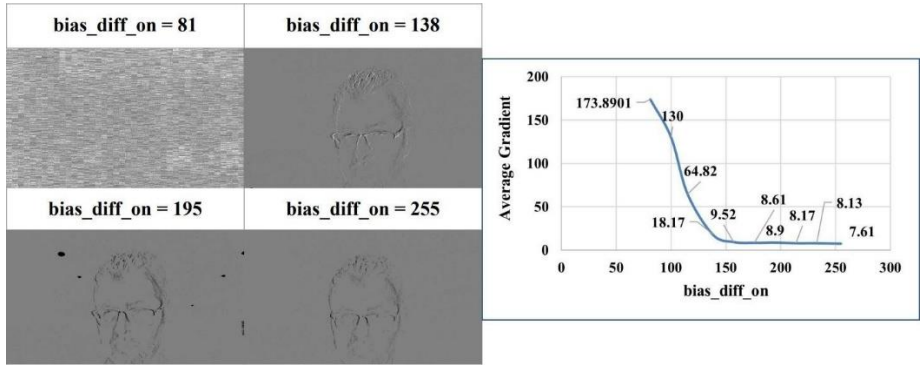

Figure 4. Left- sample output frame per each value of bias_diff_on. Right- AG Vs. bias_diff_on change.

### 4.4 Bias_diff_off

The only difference with respect to diff_on, in this case, since the sensitivity increases as we get closer to the reference amount, for higher bias values, higher AG amounts are expected. Figure 5 shows that the number of edges is increasing as bias is increased. The ideal value for this bias is about 60.

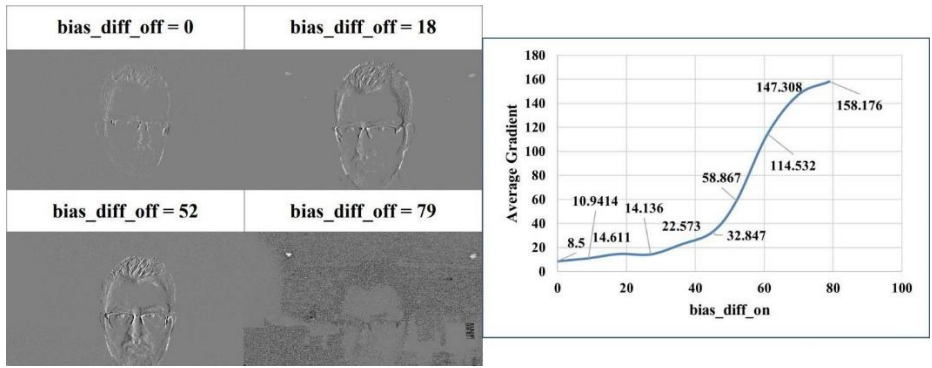

Figure 5. Left- sample output frame per each value of bias_diff_off. Right- AG Vs. bias_diff_off change.

### 4.5 Bias_refr

In contrast to the previous biases, the changes in the AG here are not linear and the curve is decreasing and increasing randomly per each value of the bias_refr. Although the number of generated events decrease when the refractory time is increased, the number of edges doesn't align with the event decrement here, and it shows completely independent behavior. Figure 6 depicts the behavior of AG with bias_refr change. Here the range between 50-100 is ideal.

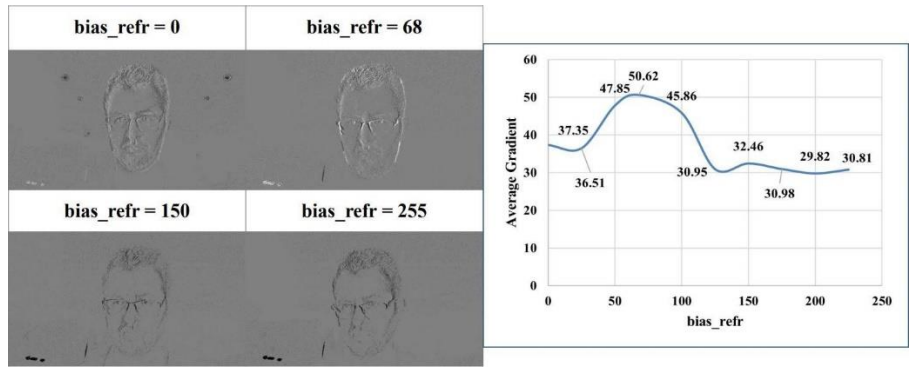

Figure 6. Left-sample output frame per each value of bias_refr. Right- AG Vs. bias_refr change.

# 5. CONCLUSION

The optimal value for each bias is the value with the highest amount of AG. For all the biases the default value is very close to the value corresponding to the maximum AG and it proves that converting the event stream to frames and measuring the AG is a good scheme to test the event stream quality. Moreover, with the curves provided per each bias, it is possible now to decide the suitable range of biases for any application. Different applications utilizing event cameras outputs require different sensor settings and since these changes highly affect the output stream, a mathematical metric is required to measure the output quality in different bias situations. We can look at the event camera as an edge detector, we studied the effect of bias change on the number of edges in event output stream. First, we converted the stream to understandable frames, then we measured the sharpness of these frames using image gradient magnitude. Except for the refractory time of the sensor which had no effect on the edges in the output AG curve, the other settings affected it linearly. Proposing a new metric specific to event streams without converting them to frames is our next goal.